\documentclass[review]{elsarticle}
\graphicspath{ {./} }
\usepackage[colorlinks, linkcolor=red, anchorcolor=blue, citecolor=green]{hyperref}
\usepackage{float}
\usepackage{verbatim} 
\usepackage{apalike}
\usepackage{makecell}
\usepackage{longtable}
\usepackage{mathtools}
\usepackage[ruled,vlined]{algorithm2e}
\usepackage{booktabs}
\usepackage{multirow}
\usepackage[normalem]{ulem}
\useunder{\uline}{\ul}{}
\usepackage{amssymb}
\usepackage{pdflscape}
\usepackage{threeparttable}
\usepackage{graphicx}
\usepackage{geometry}
\journal{Expert Systems with Applications}

\biboptions{authoryear}

\begin{document}
\begin{frontmatter}

\begin{titlepage}
\begin{center}
\vspace*{1cm}

\textbf{ \large A comparative study on machine learning models combining with outlier detection and balanced sampling methods for credit scoring}

\vspace{1.5cm}

Hongyi Qian$^{a}$ (qianhongyi@buaa.edu.cn), Shen Zhang$^{b}$ (zhangshen95726@buaa.edu.cn), Baohui Wang$^{b}$ (wangbh@buaa.edu.cn), Lei Peng$^{c}$ (penglei@chamc.com.cn), Songfeng Gao$^{c}$ (gaosongfeng@chamc.com.cn), You Song$^{b,a}$ (songyou@buaa.edu.cn) \\

\hspace{10pt}

\begin{flushleft}
\small
$^a$ School of Computer Science and Engineering, Beihang University, Beijing 100191, PR China \\
$^b$ School of Software, Beihang University, Beijing 100191, PR China \\
$^c$ HuaRong RongTong (Beijing) Technology Co., Ltd, Beijing 100033, PR China

\vspace{1cm}
\textbf{Corresponding Author:} \\
You Song \\
School of Computer Science and Engineering, Beihang University, Beijing 100191, PR China \\
School of Software, Beihang University, Beijing 100191, PR China \\
Email: songyou@buaa.edu.cn

\end{flushleft}
\end{center}
\end{titlepage}

\title{A comparative study on machine learning models combining with outlier detection and balanced sampling methods for credit scoring}

\author[label1]{Hongyi Qian}
\ead{qianhongyi@buaa.edu.cn}

\author[label2]{Shen Zhang}
\ead{zhangshen95726@buaa.edu.cn}

\author[label2]{Baohui Wang}
\ead{wangbh@buaa.edu.cn}

\author[label3]{Lei Peng}
\ead{penglei@chamc.com.cn}

\author[label3]{Songfeng Gao}
\ead{gaosongfeng@chamc.com.cn}

\author[label2,label1]{You Song\corref{cor1}}
\ead{songyou@buaa.edu.cn}

\cortext[cor1]{Corresponding author.}
\address[label1]{School of Computer Science and Engineering, Beihang University, Beijing 100191, PR China}
\address[label2]{School of Software, Beihang University, Beijing 100191, PR China}
\address[label3]{HuaRong RongTong (Beijing) Technology Co., Ltd, Beijing 100033, PR China}

\begin{abstract}
Peer-to-peer (P2P) lending platforms have grown rapidly over the past decade as the network infrastructure has improved and the demand for personal lending has grown. Such platforms allow users to create peer-to-peer lending relationships without the help of traditional financial institutions. Assessing the borrowers' credit is crucial to reduce the default rate and benign development of P2P platforms. Building a personal credit scoring machine learning model can effectively predict whether users will repay loans on the P2P platform. And the handling of data outliers and sample imbalance problems can affect the final effect of machine learning models. There have been some studies on balanced sampling methods, but the effect of outlier detection methods and their combination with balanced sampling methods on the effectiveness of machine learning models has not been fully studied. In this paper, the influence of using different outlier detection methods and balanced sampling methods on commonly used machine learning models is investigated. Experiments on 44,487 Lending Club samples show that proper outlier detection can improve the effectiveness of the machine learning model, and the balanced sampling method only has a good effect on a few machine learning models, such as MLP.
\end{abstract}


\end{frontmatter}

\section{Introduction}
\label{introduction}

Personal credit is one of the main businesses of digital financial services, and peer-to-peer (P2P) lending is the main battlefield of personal credit business. As intermediary institutions, P2P lending platforms connect borrowers and lenders, saving the cumbersome process of traditional financial institutions and reducing transaction costs \citep{Zhao2017}. This type of lending usually involves smaller amounts but comes with a higher risk of default. Since the outbreak of COVID-19 in late 2019, the global P2P lending market has had to face more and more default loans problems, which puts higher demand on lenders to review personal loan applications.

To reduce the loss of loan default, some scholars proposed methods of predicting the default probability of personal loans, which is also called credit scoring \citep{Crook2007, Dastile2020, Lessmann2015}. Specifically, credit scoring comprehensively examines the various indicators of the borrower and conducts a comprehensive assessment of their ability to fulfill economic commitments, which is usually modeled as a binary problem. For example, Lending Club is the largest online loan marketplace and borrowers can easily access lower interest rate loans through a fast online interface. When building a credit scoring model using the Lending Club dataset, the borrowers' default status is represented by the ``loan\_status'' indicator. The ``Fully Paid'' status represents that the applicant has fully paid the loan which is good, while the ``Charged Off'' status represents that the applicant has not paid the installments in due time for a long period of time which is bad.

\begin{table}[H]\footnotesize
    \centering
    \caption{Previous experiences on lending club.}
    \label{table: Lending club}
    \begin{tabular}{lp{3cm}<{\raggedright}lp{6.5cm}<{\raggedright}p{3cm}<{\raggedleft}}
    \hline
    Year & Study & Sample size & Methods & Best result with... \\ \hline
    2015 & \citep{Malekipirbazari2015} & 68,000 & LR, RF, KNN, SVM & RF \\
    2017 & \citep{Kim2017} & 332,844 & DT, Label propagation, TSVM & DT+Label propagation+TSVM \\
    2017 & \citep{Xia2017} & 49,795 & LR, RF, CSXGBoost & CSXGBoost \\
    2018 & \citep{Ma2018} & 480,018 & XGBoost, LightGBM & LightGBM \\
    2018 & \citep{Namvar2018} & 66,376 & LR, LDA, RF & LR \\
    2018 & \citep{Xia2018} & 2,642 & LR, GPC, SVM, DT, Bag-SVM, Bag-GPC, RF, XGBoost, Bstacking & Bstacking \\
    2018 & \citep{He2018} & 95,633 & LR, RF, LDA, SVM, DT, KNN, GBDT, AdaBoost, XGBoost & XGBoost \\
    2019 & \citep{Ruyu2019} & 2,022,302 & LR, Naive Bayes, DT, SVM & DT \\
    2019 & \citep{Bastani2019} & 96,200 & RF, SVM, GBDT, WDP & WDP \\
    2020 & \citep{MeloJunior2020} & 23,677 & LR, RF, BRF, MLP, SVM, SVM, XGBoost, KNU, RMkNN & BRF+KNU+RMkNN \\
    2020 & \citep{Niu2020} & 366,466 & LR, RF, DT, AdaBoost, Bagging, XGBoost, DTE-ROS, DTE-SMOTE, DTE-RUS, DTE-UnderBagging, DTE-SBD, DTE-SBC, REMDD & REMDD \\
    2020 & \citep{Xia2020} & 64,139 & LR, RF, DT, MLP, GBDT, XGBoost, CatBoost & CatBoost \\
    2020 & \citep{Song2020} & 70,860 & LR, RF, DT, MLP, GBDT, MV–ACME, AdaBoost, DM–ACME & DM–ACME \\
    2020 & \citep{Teply2020} & 212,280 & LR , MLP, LDA, SVM, RF, B-Net, Naive Bayes, DT, KNN & LR \\
    2021 & \citep{Liu2021} & 2,642 & LR, RF, DT, SVM, AdaBoost, AugBoost, GBDT, RP, MLP, Mg-GBDT & Mg-GBDT \\
    2021 & \citep{Lee2021} & 44,092 & SVM, RF, XGBoost, MLP, GCN & GCN \\
    2021 & \citep{Engelmann2021} & 42,506 & LR, RF, GBDT, KNN, DT & LR \\
    2021 & \citep{Moscato2021} & 877,956 & LR, RF, MLP & RF \\ \hline
    \end{tabular}
    \begin{tablenotes}
    \item TSVM is the transductive SVM; CSXGBoost is the cost-sensitive extreme gradient boosting model; GPC is the Gaussian process classifier; Bag-SVM is the bagging + SVM model; Bag-GPC is the bagging + GPC model; Bstacking is the bagging + stacking model; WDP is the wide and deep learning; DTE is the decision tree ensemble model; SBD is a combination of SMOTE and bagging with differentiated sampling rates; SBC is an under-sampling based on clustering; REMDD is a resampling ensemble model based on data distribution; DM–ACME is a distance-to-model and adaptive clustering-based multi-view ensemble learning method; MV–ACME is DM-ACME with ensemble strategies composed of hard probability and majority voting; BRF is the balanced random forest; KNU is the k-Nearest Oracles-Union; RMkNN is the reduced minority kNN; B-Net is the bayesian network; RP is the random projection; Mg-GBDT is the multi-grained augmented gradient boosting decision trees. AugBoost if the unsupervised feature augmentation Boosting. GCN is the graph convolutional network.
    \end{tablenotes}
\end{table}

As one of the most commonly used datasets in the field of credit scoring, many scholars have done their researches on the Lending club dataset. Some of these results are shown in table \ref{table: Lending club}. It should be noted that although the table shows the optimal results of these studies, they are not completely equivalent in comparison due to different starting conditions. However, an observable trend is that ensemble learning methods and hybrid models often achieve better results than single classifiers \citep{Liu2021, Xia2018}. Recently, some studies have applied deep learning methods to reflect nonlinear relationships between borrower’s attributes and default risk. \citep{Bastani2019} and \citep{Lee2021} applied wide and deep learning (WDP)-based and graph convolutional network (GCN)-based credit default prediction model to P2P lending respectively.

Similar to most personal credit scoring datasets, the Lending Club dataset faces the class imbalance problem. The actual default of customers is only a minority, which will affect the construction of credit scoring model. Many balanced sampling methods have been applied to solve the problem \citep{Moscato2021, Namvar2018}. Commonly used methods include simple random sampling methods such as random undersampling (RUS) and random oversampling (ROS), and more complex oversampling methods such as synthetic minority oversampling technique (SMOTE). Recently, researchers are also making other attempts to reduce the impact of sample imbalance while preserving the information of the majority class. \citep{MeloJunior2020} used a novel approach defining the local regions of a dynamic selection technique to deal with the imbalanced nature of the credit scoring dataset. \citep{Engelmann2021} applied a conditional Wasserstein GAN to construct synthetic data, which achieves better results than traditional sampling methods.

Besides the problem of the unbalanced dataset, when using real-world data to build a credit scoring model, outliers also affect the robustness of the model. And the outlier detection algorithm can detect noise or abnormal values and reduce their impact on the model. There have been some attempts at outlier detection studies on other datasets of credit scoring \citep{WeiYZZ2019, Xia2019, Zhang2021}. However, the existing Lending Club researches only focus on the balanced sampling method, while the research on outlier detection method is still in a blank state.

This article aims to build a framework for credit scoring that covering a variety of outlier detection and balanced sampling methods and models, find the applicable conditions of different methods, and find their optimal combination. To the best of our knowledge, this is the first comparative study that combines outlier detection, imbalanced data processing, and machine learning models on the Lending club dataset.

The rest of the paper is organized as follows. \hyperref[related work]{Section 2} introduces common outlier detection method, balanced sampling method, and machine learning models and their application in the field of credit scoring. \hyperref[methodology]{Section 3} presents the proposed credit scoring framework. \hyperref[experimental setup]{Section 4} shows the experimental settings, including dataset description and evaluation criteria. The experimental results are analyzed in \hyperref[experimental results and analysis]{Section 5}. \hyperref[conclusions]{Section 6} presents the conclusion and discusses the direction of future work.

\section{Related work}
\label{related work}

The proposed framework includes three technologies: outlier detection method, balanced sampling method, and machine learning model. This section will review the representative work in each area.

\subsection{Outlier detection method}

Real-world datasets usually contain a small number of outliers, because natural or man-made reasons such as small probability events occur or input errors in the data acquisition system, and these outliers will affect the performance of the machine learning model. In response to this problem, scholars have proposed a series of outlier detection algorithms to identify the outliers in the datasets. Both local outliers factor (LOF) \citep{Breunig2000} and isolation forest (IForest) \citep{Liu2008} are the classical outlier detection methods. LOF judges whether an instance is an outlier by comparing the density of each point and its neighboring points. A low-density point is more likely to be recognized as an outlier. However, \citep{Liu2008} pointed out that density is not always a good measure of outlier detection, because the density of a group of outliers is often very high, while the density of the edges of inliers points is low. IForest is ensembled by multiple decision trees, divides the hyperplane, and calculates the number of hyperplanes that isolate a sample. The number of hyperplanes that isolate an outlier in the low-density space is less than inliers. \citep{HE2003} improved LOF and proposed a cluster-based LOF outlier detection method (CBLOF). For the latest outlier detection method, \citep{Li2020} proposed a copula-based outlier detection method (COPOD) inspired by copulas for modeling multivariate data distribution. COPOD first constructs an empirical copula and then uses it to predict tail probabilities of each given data point to determine its level of extremeness, which makes it an ideal choice for high dimensional datasets.

Outlier detection algorithms have also been applied in credit scoring. \citep{Zhang2020} proposed a bagging-based local outlier factor algorithm to identify the outliers and subsequently boost them back into the training set to form the outlier-adapted training set to enhance the outlier adaptability of base classifiers. In another of their studies, \citep{Zhang2021} proposed a new voting-based outlier detection method to enhance the classic outlier detection algorithms by integrating the outlier scores through the weighted voting mechanism, and boost the outlier scores into the training set to form an outlier-adapted training set. Outlier detection is also used to reject inference in credit scoring, employing the rejected data can mitigate the sample bias caused by building the model only with the accepted applicants. For example, \citep{Xia2019} applies IForest to find ``good'' applicants whose loan applications are rejected due to accidental factors.

\subsection{Balanced sampling method}

Real-world datasets often face the class imbalance problem, the default instances in personal credit scoring datasets usually are a minority class. When the ratio of the majority sample to the minority sample exceeds 3:1, the prediction results of the classifiers will be skewed towards the majority samples. Using the balanced sampling method, a dataset with balanced positive and negative samples can be obtained to eliminate the bias in the training of the classifier.

The existing balanced sampling methods can be divided into three categories:

\begin{itemize}
    \item Under-sampling method: selecting samples from the original set to make the number of positive and negative instances equal. For example, random undersampling (RUS) is a fast and easy way to balance the data by randomly selecting a subset of data for the targeted classes. \citep{Wilson1972} proposed a method called Edited nearest neighbor (ENN) which applies a nearest-neighbors algorithm and edits the dataset by removing samples that do not agree enough with their neighborhood. In addition, instance hardness threshold (IHT) \citep{Smith2014} s a specific algorithm in which a classifier is trained on the data and the samples with lower probabilities are removed.
    \item Over-sampling method: generating new samples in the classes which are under-represented. Random oversampling (ROS) is the most naive strategy to generate new samples by randomly sampling with the replacement of the currently available samples. Apart from the ROS, there are two popular methods to over-sample minority classes: the Synthetic Minority Oversampling Technique (SMOTE) \citep{Chawla2002} and the Adaptive Synthetic (ADASYN) \citep{He2008} sampling method. SMOTE and ADASYN generate new samples by interpolation, and the samples used to interpolate new synthetic samples differ. Specifically, ADASYN focuses on generating samples next to the original samples which are wrongly classified using a k-Nearest Neighbors classifier while the basic implementation of SMOTE will not make any distinction between easy and hard samples to be classified using the nearest neighbors rule.
    \item Hybrid method: combining with over-sampling and under-sampling method. SMOTE can generate noisy samples by interpolating new points between marginal outliers and inliers. This issue can be solved by cleaning the space resulting from over-sampling. In this regard, SMOTETomek \citep{Batista2004} and SMOTEENN \citep{Batista2003} added Tomek’s link \citep{Tomek1976} and ENN to the pipeline after applying SMOTE over-sampling to obtain a cleaner space respectively.
\end{itemize}

The balanced sampling method is widely used in the credit scoring field. \citep{Moscato2021} and \citep{Namvar2018} compared the performance of various balanced sampling methods on the Lending Club dataset. In their research, RUS achieved the best results among all sampling methods. However, neither of them added the no sampling method to the experiments as a comparison. In addition to the usual methods, there have been other attempts. \citep{Engelmann2021} proposed an approach based on a conditional Wasserstein GAN that can effectively model tabular datasets with numerical and categorical variables and pays special attention to the downstream classification task through an auxiliary classifier loss.

\subsection{Machine learning model}

Machine learning is dedicated to the study of how to use the experience to improve the performance of the system itself by computing data and is an effective method to solve complex real-world problems. During the years of its development, many machine learning models have emerged. For linear models, logistic regression (LR) is the most commonly used algorithm and is often used as a benchmark model for model performance comparison \citep{AlaRaj2016}. For datasets with fewer data dimensions (e.g. Australia, Germany, Japan, etc. credit scoring datasets), it can effectively mine the interrelationships between credit data variables \citep{Abelian2017, Crook2007, Zhang2019}. However, its effectiveness is limited when dealing with high dimension, sparse data. Another commonly used credit scoring benchmark method is the multilayer perceptron (MLP) \citep{GarciaMS2019, Hajek2011, Moscato2021}. MLP is composed of an input layer, hidden layers, and an output layer, and a backpropagation algorithm is used to learn model parameters. MLP is susceptible to outliers and imbalance sample problems.

With the development of machine learning, ensemble learning models have gradually become a popular research direction. Unlike traditional learning methods for training one classifier, ensemble learning methods usually consist of multiple base classifiers. \citep{Hansen1990} and \citep{Schapire1990} pointed out that the ensemble of a set of classifiers often predicts more accurately than the best individual classifier. The representative ideas of ensemble learning are Bagging \citep{Breiman1996} and Boosting \citep{Schapire99}. Random forest (RF) \citep{Breiman2001} is one of the representatives of Bagging, which is an ensemble by multiple decision trees, and each decision tree acts as a weak learner. Gradient boosting decision tree (GBDT) is one of the popular boosting algorithms \citep{Friedman2001}. GBDT uses an additive model which is a linear combination of basis functions, and continuously reduces the residuals generated during the training process.

The ensemble tree models are usually more robust to outliers and sample imbalance problems. In credit scoring, \citep{Malekipirbazari2015} proposed an RF-based classification method for predicting borrower status. Compared with other algorithms such as LR, SVM, and KNN, RF obtains the best results. Recently, more and more state-of-art solutions are built based on GBDT. \citep{Liu2021} proposed a step-wise multi-grained augmented gradient boosting decision trees (mg-GBDT) for credit scoring. This method adopts multi-granularity scanning for feature enhancement, which enriches the input features of GBDT.

\section{The proposed methodology}
\label{methodology}
\begin{figure}[H]
    \centering
    \includegraphics[width=10cm]{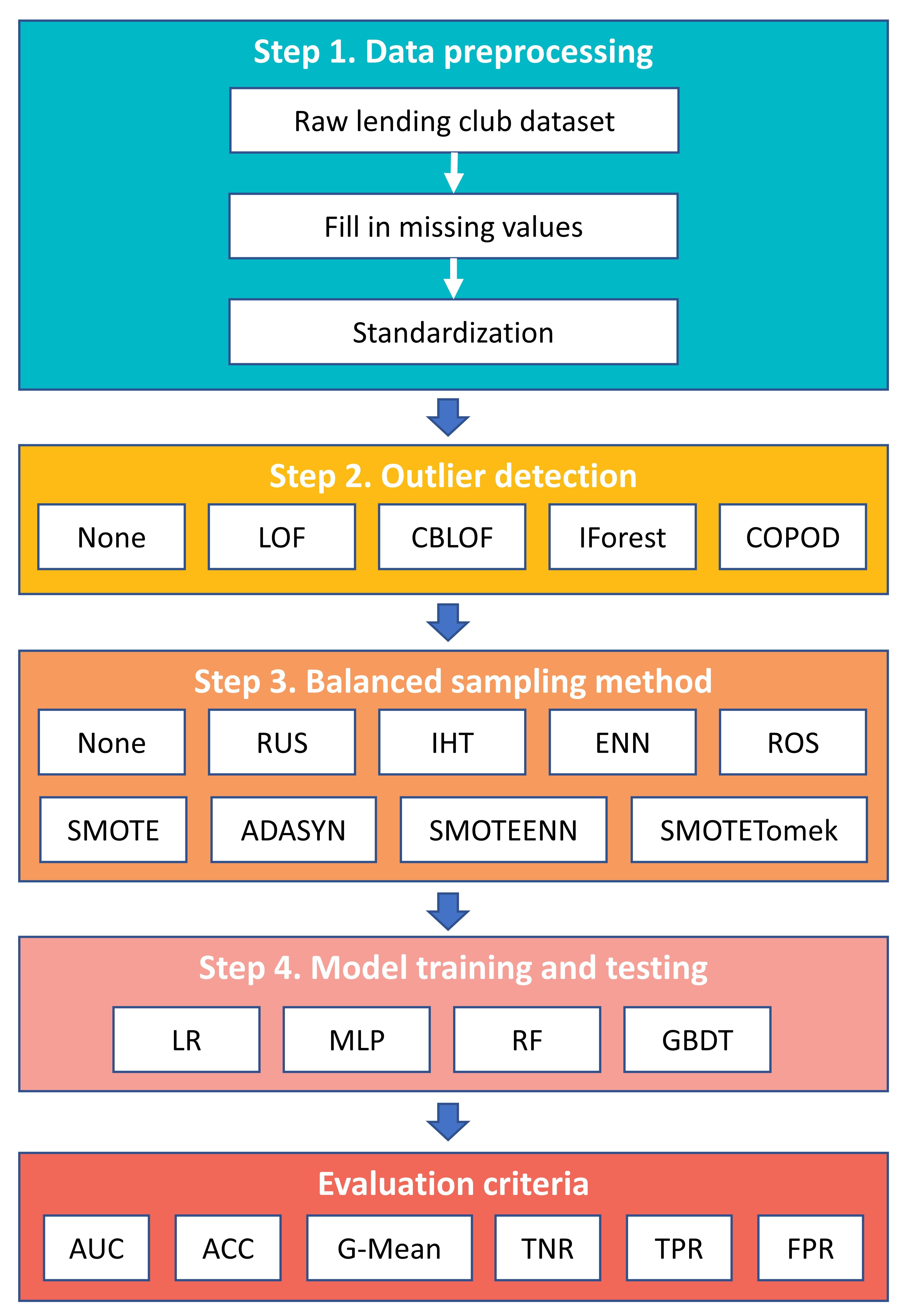}
    \caption{Flowchart of the proposed methodology.}
    \label{fig: flowchart}
\end{figure}

In this paper, a systematic study combining multiple outlier detection, balanced sampling methods, and machine learning models is proposed. The experimental process is shown in Figure \ref{fig: flowchart}. The process can be divided into four steps: data preprocessing, outlier detection, balanced sampling method, model training, and testing. Data preprocessing includes missing value filling and standardization. Then, samples containing outliers in the dataset were removed by an outlier detection algorithm. After that, the balanced sampling method is used to balance the number of positive and negative samples. Finally, the machine learning model is trained and tested on the processed dataset. The entire process is discussed in detail below.

\textbf{Step 1.} Data pre-processing

\textit{Filling missing values and standardization.}

Some numeric variables in the original dataset have missing values, which need to be preprocessed. These missing values are filled in by the mean of that indicator across all samples. In addition, since different variables usually have different value ranges, the value range of each indicator needs to be standardized to ensure the validity of the credit scoring model. In this paper, each variable is normalized according to formula \ref{equ_1}:

\begin{equation} \label{equ_1}
    x^{'}=\frac{x-u}{s}
\end{equation}

where u is the mean of the samples, and s is the standard deviation of the samples. Specifically, the scikit-learn\footnote{\url{https://scikit-learn.org}} library is used to fill in missing values and normalize the data.

\textbf{Step 2.} Outlier detection

\textit{Calculate the anomaly degree of each sample and remove samples with high anomaly degrees from the training data according to the set threshold.}

There are always some outlier samples in real datasets, which can mislead the training of machine learning models, especially linear models such as LR. In this paper, four outlier detection algorithms are used to deal with outliers, namely LOF, CBLOF, IForest, and COPOD. Specifically, the algorithms imported from pyod\footnote{\url{https://github.com/yzhao062/pyod}} library are used to evaluate the degree of outliers of the samples, and default parameters are used for each outlier detection algorithm. The samples with the top 0.5\% of the outlier scores obtained by the algorithms were removed from the training data.

\textbf{Step 3.} Balanced sampling method

\textit{Construct a positive and negative sample-balanced dataset using the balanced sampling method.}

Credit scoring datasets often face the problem of sample imbalance, which can make the classification model heavily biased. For example, MLP calculates the loss of each sample and then back-propagates the updated weights, and the sample imbalance can easily lead to the MLP model whose weight coefficients "serve" the large-class samples and have less ability to discriminate among the small-class samples.

Therefore, the balanced sampling method is used in this paper to build positive and negative sample balanced datasets, as shown in table \ref{table: sampling methods}. A total of eight sampling methods in three categories are used to solve the sample imbalance problem, among which RUS, IHT, and ENN belong to under-sampling methods, ROS, SMOTE, and ADASYN belong to over-sampling methods, SMOTEENN and SMOTETomek belong to hybrid methods. Specifically, the imbalanced-learn\footnote{\url{https://imbalanced-learn.org/}} library is used for the positive and negative sample balancing process of the samples, and default parameters are used for each algorithm.

\begin{table}[H]\footnotesize
    \centering
    \caption{Sampling methods.}
    \label{table: sampling methods}
    \begin{tabular}{lll}
    \hline
    Sampling type & Sampling methodology & Definition \\ \hline
    Under-sampling method & RUS & Random Under Sampler \\
     & IHT & Instance Hardness Threshold \\
     & ENN & Edited Nearest Neighbours \\ \hline
    Over-sampling method & ROS & Random Over Sampler \\
     & SMOTE & Synthetic Minority Oversampling Technique \\
     & ADASYN & Adaptive Synthetic \\ \hline
    Hybrid method & SMOTEENN & SMOTE + ENN \\
     & SMOTETomek & SMOTE + Tomek Links  \\ \hline
    \end{tabular}
\end{table}

\textbf{Step 4.} Model training and testing

\textit{Train the model and compare the predicted result with the real value.}

Experiments were conducted using a variety of machine learning models, including LR, MLP, RF, and GBDT. Specifically, the machine learning model is imported from the scikit-learn library, and the 5-fold cross-validation training process is used. Each time 80\% of the data is selected as the training set, and the remaining 20\% is used as the test set. The hyperparameters of each model are shown in table \ref{table: parameters}.

\begin{table}[H]\footnotesize
    \centering
    \caption{Parameters of different models.}
    \label{table: parameters}
    \begin{tabular}{lll}
    \hline
    Classifiers & Parameters & Values \\ \hline
    LR & penalty & l2 \\
     & tol & 1e-4 \\
     & solver & lbfgs \\ \hline
    MLP & hidden\_layer\_sizes & (100, ) \\
     & activation & relu \\
     & learning\_rate\_init & 0.01 \\
     & early\_stopping & TRUE \\ \hline
    RF & n\_estimators & 100 \\
     & max\_depth & 4 \\ \hline
    GBDT & n\_estimators & 100 \\
     & learning\_rate & 0.1 \\
     & max\_depth & 4 \\ \hline
    \end{tabular}
\end{table}

\section{Experimental setup}
\label{experimental setup}
This section describes the experimental setup in detail, consisting of two parts, dataset description, and evaluation criteria.

\subsection{Dataset description}
The dataset used in this paper is from Lending Club\footnote{\url{https://www.lendingclub.com/}} in Q2 2017. Lending Club is a US P2P lending company, headquartered in San Francisco, California. It was the first P2P lender to offer loan trading on a secondary market and is the world's largest P2P lending platform. The original dataset contains a total of 105,451 samples and 151 features, with ``loan\_status'' as the target variable. As shown in table \ref{table: Dataset characterization}, ``loan\_status'' has 7 states, ``Current'', ``Fully Paid'', ``Charged Off'', ``Late (31-120 days)'', ``In Grace Period'', ``Late (16-30 days)'', and ``Default''. Referring to the practice in previous papers \citep{Moscato2021, Namvar2018}, only the samples with ``loan\_status'' status as ``Charged off'' and ``Fully Paid'' are taken as positive and negative samples respectively. This results in an unbalanced dataset containing 44,487 samples with a positive sample ratio of 23.80\%.

\begin{table}[H]\footnotesize
    \centering
    \caption{Dataset characterization.}
    \label{table: Dataset characterization}
    \begin{tabular}{ll}
    \hline
    Loan status & Amount \\ \hline
    Current & 58201 \\
    Fully Paid & 33901 \\
    Charged Off & 10586 \\
    Late (31-120 days) & 1782 \\
    In Grace Period & 650 \\
    Late (16-30 days) & 326 \\
    Default & 5 \\
    \textbf{Total} & 105451 \\ \hline
    \end{tabular}
\end{table}

Variables with a high proportion of missing values in the original features were not used. Of the remaining variables, only those that would be available to an investor before deciding to fund the loan were chosen. As a result, except the target variable ``loan\_status'', 24 variables are actually used. The descriptions of each field were shown in table \ref{table: Indicator descriptions}.

\begin{table}[H]\footnotesize
    \centering
    \caption{Indicator descriptions.}
    \label{table: Indicator descriptions}
    \begin{tabular}{ll}
    \hline
    LoanStatNew & Description \\ \hline
    addr\_state & The state provided by the borrower in the loan application. \\
    annual\_inc & The self-reported annual income provided by the borrower during registration. \\
    application\_type & Indicates whether the loan is an individual application or a joint application with two co-borrowers. \\
    dti & Borrower’s total monthly debt payments divided by the borrower’s self-reported monthly income. \\
    earliest\_cr\_line & The month the borrower's earliest reported credit line was opened. \\
    emp\_length & Employment length in years. \\
    fico\_range\_high & The upper boundary range the borrower’s FICO at loan origination belongs to. \\
    fico\_range\_low & The lower boundary range the borrower’s FICO at loan origination belongs to. \\
    home\_ownership & The home ownership status provided by the borrower during registration. \\
    initial\_list\_status & The initial listing status of the loan. \\
    installment & The monthly payment owed by the borrower if the loan originates. \\
    int\_rate & Interest Rate on the loan. \\
    loan\_amnt & The listed amount of the loan applied for by the borrower. \\
    loan\_status & Current status of the loan. \\
    mort\_acc & Number of mortgage accounts. \\
    open\_acc & The number of open credit lines in the borrower's credit file. \\
    pub\_rec & Number of derogatory public records. \\
    pub\_rec\_bankruptcies & Number of public record bankruptcies. \\
    purpose & A category provided by the borrower for the loan request. \\
    revol\_bal & Total credit revolving balance. \\
    revol\_util & Revolving line utilization rate, or the amount of credit relative to all available revolving credit. \\
    sub\_grade & LC assigned loan subgrade. \\
    term & The number of payments on the loan. Values are in months and can be either 36 or 60. \\
    total\_acc & The total number of credit lines currently in the borrower's credit file. \\
    verification\_status & Indicates if income was verified by LC, not verified, or if the income source was verified. \\ \hline
    \end{tabular}
\end{table}

Some variables are preprocessed before input to the model:
\begin{itemize}
    \item Models such as LR cannot directly deal with class-type variables. The category variables ``sub\_grade'', ``home\_ownership'', ``verification\_status'', ``purpose'', ``addr\_state'', ``initial\_list\_status'', and ``application\_type'' are coded as one-hot;
    \item The original credit score ``fico\_score'' provides the values ``low'' and ``high'', which are replaced by their average value;
    \item Log transformations are performed on exponential numerical variables such as ``annual\_inc'' and ``revol\_bal'';
\end{itemize}

After the above processing, the actual number of variables input to the model is 123. Table \ref{table: Indicators statistics} shows the statistical characteristics of the numerical variables.

\begin{table}[H]\footnotesize
    \centering
    \caption{Indicators statistics.}
    \label{table: Indicators statistics}
    \begin{tabular}{lllllllll}
    \hline
    variable name & count & mean & std & min & 25\% & 50\% & 75\% & max \\ \hline
    loan\_amnt & 44487 & 14147.24 & 9343.50 & 1000.00 & 7000.00 & 12000.00 & 20000.00 & 40000.00 \\
    term & 44487 & 41.59 & 10.15 & 36.00 & 36.00 & 36.00 & 36.00 & 60.00 \\
    int\_rate & 44487 & 13.90 & 5.58 & 5.32 & 10.42 & 12.74 & 16.02 & 30.99 \\
    installment & 44487 & 436.24 & 287.29 & 30.12 & 225.65 & 356.78 & 583.89 & 1719.83 \\
    emp\_length & 41614 & 5.88 & 3.76 & 0.00 & 2.00 & 6.00 & 10.00 & 10.00 \\
    dti & 44456 & 18.68 & 13.78 & 0.00 & 11.95 & 17.87 & 24.34 & 999.00 \\
    earliest\_cr\_line & 44487 & 2000.65 & 7.59 & 1957.00 & 1997.00 & 2002.00 & 2006.00 & 2014.00 \\
    open\_acc & 44487 & 11.85 & 5.74 & 0.00 & 8.00 & 11.00 & 15.00 & 88.00 \\
    pub\_rec & 44487 & 0.28 & 0.66 & 0.00 & 0.00 & 0.00 & 0.00 & 22.00 \\
    revol\_util & 44455 & 46.75 & 24.53 & 0.00 & 27.70 & 45.90 & 65.10 & 138.90 \\
    total\_acc & 44487 & 24.48 & 12.11 & 2.00 & 16.00 & 23.00 & 31.00 & 146.00 \\
    mort\_acc & 44487 & 1.51 & 1.80 & 0.00 & 0.00 & 1.00 & 2.00 & 18.00 \\
    pub\_rec\_bankruptcies & 44487 & 0.16 & 0.41 & 0.00 & 0.00 & 0.00 & 0.00 & 6.00 \\
    log\_annual\_inc & 44487 & 4.83 & 0.27 & 0.00 & 4.68 & 4.83 & 4.98 & 6.75 \\
    fico\_score & 44487 & 699.73 & 33.49 & 662.00 & 672.00 & 692.00 & 717.00 & 847.50 \\
    log\_revol\_bal & 44487 & 3.96 & 0.55 & 0.00 & 3.73 & 4.02 & 4.27 & 6.02 \\ \hline
    \end{tabular}
    \end{table}

\subsection{Evaluation criteria}
Comparing the effectiveness of models requires the use of appropriate evaluation criteria. In the field of credit scoring, overall accuracy (ACC) is one of the most commonly used indicators, which is defined as the ratio of the number of correctly classified samples to the total number of samples. In addition, TNR, TPR, FPR, and G-mean were also popular evaluation indicators. These indicators can be calculated according to formulas \ref{equ_2} - \ref{equ_6}. The probability threshold for classifying the test set is set to exactly the extent that its positive and negative sample ratios remain consistent with the original dataset.

\begin{equation} \label{equ_2}
    ACC = \frac{TP+FN}{TP+FP+FN+TN}
\end{equation}
\begin{equation} \label{equ_3}
    TNR = \frac{TN}{TN+FP}
\end{equation}
\begin{equation} \label{equ_4}
    TPR = \frac{TP}{TP+FN}
\end{equation}
\begin{equation} \label{equ_5}
    FPR = \frac{FP}{FP+TN}
\end{equation}
\begin{equation} \label{equ_6}
    G-mean = \sqrt{TNR*TPR}
\end{equation}

The confusion matrix consisting of TP, TN, FP, and FN in these formulas is shown in table \ref{table: confusion matrix}, where TP and TN represent the numbers of correctly classified normal and default users, respectively. FP and FN denote the numbers of misclassified normal and default users, respectively.

\begin{table}[H]\footnotesize
    \centering
    \caption{The confusion matrix.}
    \label{table: confusion matrix}
    \begin{tabular}{lll}
    \hline
    Actual Class                                                                  & Prediction Results &    \\ \cline{2-3}
     & \begin{tabular}[c]{@{}l@{}}Positive class (Default)\end{tabular} & \begin{tabular}[c]{@{}l@{}}Negative class (Normal)\end{tabular} \\ \hline
    \begin{tabular}[c]{@{}l@{}}Positive class (Default)\end{tabular} & TP                  & FN \\
    \begin{tabular}[c]{@{}l@{}}Negative class (Normal)\end{tabular}   & FP                  & TN \\ \hline
    \end{tabular}
\end{table}

However, the above metrics are strongly influenced by the probability threshold, and the comprehensive performance of the classifier can be well evaluated by using the area under the curve (AUC). AUC is an evaluation metric based on the receiver operating characteristic (ROC) curve, which is equal to the area under the ROC curve and well reflects the trade-off between the true positive rate and the false positive rate of the prediction model under different probability thresholds. The AUC will be used as the main evaluation metric.

\section{Experimental results and analysis}
\label{experimental results and analysis}

The performance of classifiers was evaluated according to different outlier detection and sampling strategies

\subsection{The original results of each model}
Table \ref{table: LR} shows the classification results of LR. In general, the difference brought by using different outlier detection and balanced sampling methods is not significant, the difference between the first and the bottom of the AUC ranking is only 0.0064. Among the outlier detection and balanced sampling methods, the most effective are LOF and None (AUC 0.7079), and the most ineffective are None and ADASYN (AUC 0.7015). The most effective and least effective criteria are the maximum and minimum AUC that can be achieved using the strategy, as well as for other models. However, under any other criteria like G-mean, the best strategy to match LR is IForest and SMOTE.

\begin{table}[H]\scriptsize
    \centering
    \caption{Classification results (LR).}
    \label{table: LR}
    \begin{tabular}{lllllllll}
    \hline
    \textbf{Model} & \textbf{reject\_strategy} & \textbf{sampling\_strategy} & \textbf{ACC} & \textbf{AUC} & \textbf{G-mean} & \textbf{TPR} & \textbf{TNR} & \textbf{FPR} \\ \hline
    LR & None & None & 0.7298 & 0.7071 & 0.5893 & 0.4203 & 0.8264 & 0.1736 \\
    LR & None & RUS & 0.7294 & 0.7066 & 0.5887 & 0.4196 & 0.8262 & 0.1738 \\
    LR & None & IHT & 0.7279 & 0.7031 & 0.5861 & 0.4163 & 0.8252 & 0.1748 \\
    LR & None & ENN & 0.7282 & 0.7040 & 0.5867 & 0.4170 & 0.8254 & 0.1746 \\
    LR & None & ROS & 0.7291 & 0.7072 & 0.5882 & 0.4189 & 0.8260 & 0.1740 \\
    LR & None & SMOTE & 0.7292 & 0.7040 & 0.5883 & 0.4190 & 0.8260 & 0.1740 \\
    LR & None & ADASYN & 0.7271 & 0.7015 & 0.5848 & 0.4147 & 0.8247 & 0.1753 \\
    LR & None & SMOTEENN & 0.7272 & 0.7015 & 0.5849 & 0.4148 & 0.8247 & 0.1753 \\
    LR & None & SMOTETomek & 0.7296 & 0.7041 & 0.5891 & 0.4200 & 0.8263 & 0.1737 \\
    LR & COPOD & None & 0.7297 & 0.7077 & 0.5893 & 0.4202 & 0.8264 & 0.1736 \\
    LR & COPOD & RUS & 0.7300 & 0.7071 & 0.5897 & 0.4208 & 0.8265 & 0.1735 \\
    LR & COPOD & IHT & 0.7279 & 0.7026 & 0.5862 & 0.4164 & 0.8252 & 0.1748 \\
    LR & COPOD & ENN & 0.7278 & 0.7050 & 0.5860 & 0.4162 & 0.8251 & 0.1749 \\
    LR & COPOD & ROS & 0.7294 & 0.7077 & 0.5887 & 0.4195 & 0.8261 & 0.1739 \\
    LR & COPOD & SMOTE & 0.7300 & 0.7042 & 0.5896 & 0.4207 & 0.8265 & 0.1735 \\
    LR & COPOD & ADASYN & 0.7272 & 0.7013 & 0.5849 & 0.4148 & 0.8247 & 0.1753 \\
    LR & COPOD & SMOTEENN & 0.7274 & 0.7025 & 0.5853 & 0.4154 & 0.8249 & 0.1751 \\
    LR & COPOD & SMOTETomek & 0.7296 & 0.7042 & 0.5890 & 0.4198 & 0.8263 & 0.1738 \\
    LR & IForest & None & 0.7303 & 0.7074 & 0.5902 & 0.4214 & 0.8267 & 0.1733 \\
    LR & IForest & RUS & 0.7283 & 0.7067 & 0.5867 & 0.4171 & 0.8254 & 0.1746 \\
    LR & IForest & IHT & 0.7280 & 0.7030 & 0.5863 & 0.4165 & 0.8252 & 0.1748 \\
    LR & IForest & ENN & 0.7282 & 0.7047 & 0.5867 & 0.4170 & 0.8254 & 0.1746 \\
    LR & IForest & ROS & 0.7293 & 0.7074 & 0.5886 & 0.4194 & 0.8261 & 0.1739 \\
    LR & IForest & SMOTE & \textbf{0.7306} & 0.7046 & \textbf{0.5907} & \textbf{0.4220} & \textbf{0.8269} & \textbf{0.1731} \\
    LR & IForest & ADASYN & 0.7278 & 0.7017 & 0.5859 & 0.4161 & 0.8251 & 0.1749 \\
    LR & IForest & SMOTEENN & 0.7273 & 0.7027 & 0.5851 & 0.4151 & 0.8248 & 0.1752 \\
    LR & IForest & SMOTETomek & 0.7302 & 0.7046 & 0.5901 & 0.4213 & 0.8267 & 0.1733 \\
    LR & CBLOF & None & 0.7298 & 0.7073 & 0.5894 & 0.4204 & 0.8264 & 0.1736 \\
    LR & CBLOF & RUS & 0.7287 & 0.7069 & 0.5876 & 0.4181 & 0.8257 & 0.1743 \\
    LR & CBLOF & IHT & 0.7288 & 0.7028 & 0.5876 & 0.4182 & 0.8257 & 0.1743 \\
    LR & CBLOF & ENN & 0.7283 & 0.7050 & 0.5867 & 0.4171 & 0.8254 & 0.1746 \\
    LR & CBLOF & ROS & 0.7296 & 0.7072 & 0.5890 & 0.4199 & 0.8263 & 0.1737 \\
    LR & CBLOF & SMOTE & 0.7288 & 0.7043 & 0.5877 & 0.4183 & 0.8258 & 0.1742 \\
    LR & CBLOF & ADASYN & 0.7265 & 0.7016 & 0.5838 & 0.4135 & 0.8243 & 0.1757 \\
    LR & CBLOF & SMOTEENN & 0.7276 & 0.7027 & 0.5856 & 0.4158 & 0.8250 & 0.1750 \\
    LR & CBLOF & SMOTETomek & 0.7287 & 0.7043 & 0.5875 & 0.4180 & 0.8257 & 0.1743 \\
    LR & LOF & None & 0.7299 & \textbf{0.7079} & 0.5895 & 0.4205 & 0.8265 & 0.1735 \\
    LR & LOF & RUS & 0.7298 & 0.7064 & 0.5894 & 0.4204 & 0.8264 & 0.1736 \\
    LR & LOF & IHT & 0.7269 & 0.7023 & 0.5844 & 0.4143 & 0.8245 & 0.1755 \\
    LR & LOF & ENN & 0.7281 & 0.7051 & 0.5864 & 0.4167 & 0.8253 & 0.1747 \\
    LR & LOF & ROS & 0.7303 & 0.7076 & 0.5903 & 0.4214 & 0.8268 & 0.1732 \\
    LR & LOF & SMOTE & 0.7298 & 0.7039 & 0.5894 & 0.4204 & 0.8264 & 0.1736 \\
    LR & LOF & ADASYN & 0.7272 & 0.7017 & 0.5850 & 0.4149 & 0.8247 & 0.1753 \\
    LR & LOF & SMOTEENN & 0.7277 & 0.7024 & 0.5858 & 0.4160 & 0.8250 & 0.1750 \\
    LR & LOF & SMOTETomek & 0.7296 & 0.7039 & 0.5890 & 0.4198 & 0.8263 & 0.1738 \\ \hline
    \end{tabular}
\end{table}

Table \ref{table: MLP} shows the classification results of MLP, the use of different outlier detection and balanced sampling methods brought the greatest differences, with the difference between the first and the last ranked AUC being 0.0879. This suggests that the MLP is very sensitive to data outliers and sample imbalances. Among the outlier detection and balanced sampling methods, the most effective were LOF and IHT (AUC 0.7041), slightly below the LR. And the least effective were None and ADASYN (AUC 0.6162), which is the same as LR. Under any other criteria like G-mean, the best strategy to match MLP is CBLOF and IHT.

\begin{table}[H]\scriptsize
    \centering
    \caption{Classification results (MLP).}
    \label{table: MLP}
    \begin{tabular}{lllllllll}
    \hline
    \textbf{Model} & \textbf{reject\_strategy} & \textbf{sampling\_strategy} & \textbf{ACC} & \textbf{AUC} & \textbf{G-mean} & \textbf{TPR} & \textbf{TNR} & \textbf{FPR} \\ \hline
    MLP & None & None & 0.7230 & 0.6953 & 0.5778 & 0.4061 & 0.8220 & 0.1780 \\
    MLP & None & RUS & 0.7225 & 0.6936 & 0.5768 & 0.4049 & 0.8216 & 0.1784 \\
    MLP & None & IHT & 0.7282 & 0.7014 & 0.5866 & 0.4169 & 0.8253 & 0.1747 \\
    MLP & None & ENN & 0.7131 & 0.6730 & 0.5606 & 0.3854 & 0.8155 & 0.1845 \\
    MLP & None & ROS & 0.6938 & 0.6267 & 0.5260 & 0.3447 & 0.8028 & 0.1972 \\
    MLP & None & SMOTE & 0.6908 & 0.6201 & 0.5204 & 0.3383 & 0.8008 & 0.1992 \\
    MLP & None & ADASYN & 0.6880 & 0.6162 & 0.5154 & 0.3325 & 0.7990 & 0.2010 \\
    MLP & None & SMOTEENN & 0.7075 & 0.6553 & 0.5506 & 0.3735 & 0.8118 & 0.1882 \\
    MLP & None & SMOTETomek & 0.6928 & 0.6245 & 0.5242 & 0.3426 & 0.8021 & 0.1979 \\
    MLP & COPOD & None & 0.7246 & 0.6979 & 0.5804 & 0.4094 & 0.8230 & 0.1770 \\
    MLP & COPOD & RUS & 0.7240 & 0.6944 & 0.5794 & 0.4081 & 0.8226 & 0.1774 \\
    MLP & COPOD & IHT & 0.7278 & 0.7029 & 0.5860 & 0.4162 & 0.8251 & 0.1749 \\
    MLP & COPOD & ENN & 0.7157 & 0.6774 & 0.5650 & 0.3907 & 0.8172 & 0.1828 \\
    MLP & COPOD & ROS & 0.6904 & 0.6257 & 0.5198 & 0.3376 & 0.8006 & 0.1994 \\
    MLP & COPOD & SMOTE & 0.6901 & 0.6176 & 0.5192 & 0.3369 & 0.8004 & 0.1996 \\
    MLP & COPOD & ADASYN & 0.6913 & 0.6227 & 0.5216 & 0.3395 & 0.8012 & 0.1988 \\
    MLP & COPOD & SMOTEENN & 0.7077 & 0.6532 & 0.5509 & 0.3738 & 0.8119 & 0.1881 \\
    MLP & COPOD & SMOTETomek & 0.6886 & 0.6240 & 0.5166 & 0.3339 & 0.7994 & 0.2006 \\
    MLP & IForest & None & 0.7212 & 0.6917 & 0.5746 & 0.4023 & 0.8208 & 0.1792 \\
    MLP & IForest & RUS & 0.7229 & 0.6960 & 0.5775 & 0.4058 & 0.8219 & 0.1781 \\
    MLP & IForest & IHT & 0.7271 & 0.7028 & 0.5847 & 0.4146 & 0.8246 & 0.1754 \\
    MLP & IForest & ENN & 0.7157 & 0.6775 & 0.5650 & 0.3907 & 0.8172 & 0.1828 \\
    MLP & IForest & ROS & 0.6904 & 0.6273 & 0.5199 & 0.3376 & 0.8006 & 0.1994 \\
    MLP & IForest & SMOTE & 0.6936 & 0.6281 & 0.5257 & 0.3444 & 0.8027 & 0.1973 \\
    MLP & IForest & ADASYN & 0.6917 & 0.6242 & 0.5221 & 0.3403 & 0.8014 & 0.1986 \\
    MLP & IForest & SMOTEENN & 0.7055 & 0.6537 & 0.5471 & 0.3693 & 0.8105 & 0.1895 \\
    MLP & IForest & SMOTETomek & 0.6945 & 0.6285 & 0.5273 & 0.3462 & 0.8032 & 0.1968 \\
    MLP & CBLOF & None & 0.7256 & 0.6980 & 0.5822 & 0.4115 & 0.8237 & 0.1763 \\
    MLP & CBLOF & RUS & 0.7223 & 0.6952 & 0.5765 & 0.4046 & 0.8215 & 0.1785 \\
    MLP & CBLOF & IHT & \textbf{0.7293} & 0.7035 & \textbf{0.5886} & \textbf{0.4194} & \textbf{0.8261} & \textbf{0.1739} \\
    MLP & CBLOF & ENN & 0.7114 & 0.6655 & 0.5574 & 0.3816 & 0.8143 & 0.1857 \\
    MLP & CBLOF & ROS & 0.6916 & 0.6302 & 0.5220 & 0.3401 & 0.8014 & 0.1986 \\
    MLP & CBLOF & SMOTE & 0.6939 & 0.6280 & 0.5261 & 0.3448 & 0.8028 & 0.1972 \\
    MLP & CBLOF & ADASYN & 0.6869 & 0.6184 & 0.5133 & 0.3302 & 0.7983 & 0.2017 \\
    MLP & CBLOF & SMOTEENN & 0.7072 & 0.6566 & 0.5500 & 0.3728 & 0.8116 & 0.1884 \\
    MLP & CBLOF & SMOTETomek & 0.6918 & 0.6235 & 0.5224 & 0.3406 & 0.8015 & 0.1985 \\
    MLP & LOF & None & 0.7266 & 0.6969 & 0.5839 & 0.4136 & 0.8243 & 0.1757 \\
    MLP & LOF & RUS & 0.7238 & 0.6945 & 0.5792 & 0.4078 & 0.8225 & 0.1775 \\
    MLP & LOF & IHT & 0.7284 & \textbf{0.7041} & 0.5870 & 0.4174 & 0.8255 & 0.1745 \\
    MLP & LOF & ENN & 0.7081 & 0.6552 & 0.5516 & 0.3747 & 0.8122 & 0.1878 \\
    MLP & LOF & ROS & 0.6955 & 0.6311 & 0.5291 & 0.3482 & 0.8039 & 0.1961 \\
    MLP & LOF & SMOTE & 0.6921 & 0.6253 & 0.5228 & 0.3410 & 0.8017 & 0.1983 \\
    MLP & LOF & ADASYN & 0.6895 & 0.6223 & 0.5182 & 0.3358 & 0.8000 & 0.2000 \\
    MLP & LOF & SMOTEENN & 0.7082 & 0.6596 & 0.5518 & 0.3750 & 0.8122 & 0.1878 \\
    MLP & LOF & SMOTETomek & 0.6922 & 0.6247 & 0.5232 & 0.3414 & 0.8018 & 0.1982 \\ \hline
    \end{tabular}
\end{table}

The classification results of RF are shown in table \ref{table: RF}, the difference between the top and bottom AUC ranking is 0.0207, the fluctuation is between LR and MLP. Among the outlier detection and balanced sampling methods, the most effective were IForest and None (AUC 0.6985), below the upper limit of what can be achieved by LR and MLP. And the least effective were IForest and ADASYN (AUC 0.6778), both the best and worst RF models use IForest. Among the sampling methods, ADASYN still gets the worst results.

\begin{table}[H]\scriptsize
    \centering
    \caption{Classification results (RF).}
    \label{table: RF}
    \begin{tabular}{lllllllll}
    \hline
    \textbf{Model} & \textbf{reject\_strategy} & \textbf{sampling\_strategy} & \textbf{ACC} & \textbf{AUC} & \textbf{G-mean} & \textbf{TPR} & \textbf{TNR} & \textbf{FPR} \\ \hline
    RF & None & None & 0.7242 & 0.6977 & 0.5797 & 0.4085 & 0.8227 & 0.1773 \\
    RF & None & RUS & 0.7247 & 0.6981 & 0.5806 & 0.4096 & 0.8231 & 0.1769 \\
    RF & None & IHT & 0.7184 & 0.6920 & 0.5697 & 0.3964 & 0.8189 & 0.1811 \\
    RF & None & ENN & 0.7229 & 0.6943 & 0.5774 & 0.4058 & 0.8219 & 0.1781 \\
    RF & None & ROS & 0.7221 & 0.6956 & 0.5761 & 0.4041 & 0.8213 & 0.1787 \\
    RF & None & SMOTE & 0.7161 & 0.6819 & 0.5657 & 0.3916 & 0.8174 & 0.1826 \\
    RF & None & ADASYN & 0.7149 & 0.6778 & 0.5637 & 0.3891 & 0.8167 & 0.1833 \\
    RF & None & SMOTEENN & 0.7153 & 0.6856 & 0.5642 & 0.3898 & 0.8169 & 0.1831 \\
    RF & None & SMOTETomek & 0.7171 & 0.6827 & 0.5675 & 0.3937 & 0.8181 & 0.1819 \\
    RF & COPOD & None & 0.7246 & 0.6976 & 0.5805 & 0.4094 & 0.8230 & 0.1770 \\
    RF & COPOD & RUS & 0.7237 & 0.6971 & 0.5789 & 0.4076 & 0.8224 & 0.1776 \\
    RF & COPOD & IHT & 0.7188 & 0.6916 & 0.5704 & 0.3973 & 0.8192 & 0.1808 \\
    RF & COPOD & ENN & 0.7222 & 0.6947 & 0.5764 & 0.4044 & 0.8214 & 0.1786 \\
    RF & COPOD & ROS & 0.7226 & 0.6969 & 0.5771 & 0.4053 & 0.8217 & 0.1783 \\
    RF & COPOD & SMOTE & 0.7176 & 0.6833 & 0.5683 & 0.3947 & 0.8184 & 0.1816 \\
    RF & COPOD & ADASYN & 0.7145 & 0.6780 & 0.5629 & 0.3881 & 0.8163 & 0.1837 \\
    RF & COPOD & SMOTEENN & 0.7166 & 0.6855 & 0.5665 & 0.3925 & 0.8177 & 0.1823 \\
    RF & COPOD & SMOTETomek & 0.7163 & 0.6827 & 0.5661 & 0.3920 & 0.8175 & 0.1825 \\
    RF & IForest & None & \textbf{0.7248} & \textbf{0.6985} & \textbf{0.5808} & \textbf{0.4098} & \textbf{0.8231} & \textbf{0.1769} \\
    RF & IForest & RUS & 0.7231 & 0.6980 & 0.5779 & 0.4063 & 0.8220 & 0.1780 \\
    RF & IForest & IHT & 0.7199 & 0.6920 & 0.5723 & 0.3995 & 0.8199 & 0.1801 \\
    RF & IForest & ENN & 0.7224 & 0.6942 & 0.5766 & 0.4047 & 0.8215 & 0.1785 \\
    RF & IForest & ROS & 0.7217 & 0.6968 & 0.5754 & 0.4033 & 0.8211 & 0.1789 \\
    RF & IForest & SMOTE & 0.7154 & 0.6817 & 0.5645 & 0.3901 & 0.8170 & 0.1830 \\
    RF & IForest & ADASYN & 0.7142 & 0.6778 & 0.5624 & 0.3875 & 0.8162 & 0.1838 \\
    RF & IForest & SMOTEENN & 0.7162 & 0.6861 & 0.5659 & 0.3918 & 0.8175 & 0.1825 \\
    RF & IForest & SMOTETomek & 0.7156 & 0.6822 & 0.5649 & 0.3906 & 0.8171 & 0.1829 \\
    RF & CBLOF & None & 0.7247 & 0.6972 & 0.5806 & 0.4095 & 0.8230 & 0.1770 \\
    RF & CBLOF & RUS & 0.7236 & 0.6967 & 0.5788 & 0.4074 & 0.8224 & 0.1776 \\
    RF & CBLOF & IHT & 0.7177 & 0.6909 & 0.5686 & 0.3950 & 0.8185 & 0.1815 \\
    RF & CBLOF & ENN & 0.7221 & 0.6936 & 0.5761 & 0.4042 & 0.8214 & 0.1786 \\
    RF & CBLOF & ROS & 0.7214 & 0.6951 & 0.5749 & 0.4026 & 0.8209 & 0.1791 \\
    RF & CBLOF & SMOTE & 0.7144 & 0.6826 & 0.5628 & 0.3880 & 0.8163 & 0.1837 \\
    RF & CBLOF & ADASYN & 0.7141 & 0.6782 & 0.5623 & 0.3874 & 0.8161 & 0.1839 \\
    RF & CBLOF & SMOTEENN & 0.7155 & 0.6848 & 0.5646 & 0.3903 & 0.8170 & 0.1830 \\
    RF & CBLOF & SMOTETomek & 0.7156 & 0.6827 & 0.5648 & 0.3905 & 0.8171 & 0.1829 \\
    RF & LOF & None & 0.7237 & 0.6970 & 0.5788 & 0.4075 & 0.8224 & 0.1776 \\
    RF & LOF & RUS & 0.7238 & 0.6970 & 0.5791 & 0.4077 & 0.8225 & 0.1775 \\
    RF & LOF & IHT & 0.7182 & 0.6925 & 0.5694 & 0.3959 & 0.8188 & 0.1812 \\
    RF & LOF & ENN & 0.7212 & 0.6940 & 0.5747 & 0.4024 & 0.8208 & 0.1792 \\
    RF & LOF & ROS & 0.7229 & 0.6961 & 0.5775 & 0.4058 & 0.8219 & 0.1781 \\
    RF & LOF & SMOTE & 0.7159 & 0.6828 & 0.5654 & 0.3912 & 0.8173 & 0.1827 \\
    RF & LOF & ADASYN & 0.7135 & 0.6781 & 0.5611 & 0.3860 & 0.8157 & 0.1843 \\
    RF & LOF & SMOTEENN & 0.7162 & 0.6853 & 0.5659 & 0.3918 & 0.8175 & 0.1825 \\
    RF & LOF & SMOTETomek & 0.7158 & 0.6823 & 0.5653 & 0.3910 & 0.8173 & 0.1827 \\ \hline
    \end{tabular}
\end{table}

The classification results of GBDT is shown in table \ref{table: GBDT}. The difference between the top and bottom AUC ranking is 0.0116, it's closer to the RF which is the same as the tree model. Among the outlier detection and balanced sampling methods, the most effective are LOF and None (AUC 0.7107), which is the same as LR and is the best of the four models. The least effective are CBLOF and ADASYN (AUC 0.6991), the worst effects of the four models all use ADASYN.

\begin{table}[H]\scriptsize
    \centering
    \caption{Classification results (GBDT).}
    \label{table: GBDT}
    \begin{tabular}{lllllllll}
    \hline
    \textbf{Model} & \textbf{reject\_strategy} & \textbf{sampling\_strategy} & \textbf{ACC} & \textbf{AUC} & \textbf{G-mean} & \textbf{TPR} & \textbf{TNR} & \textbf{FPR} \\ \hline
    GBDT & None & None & 0.7319 & 0.7102 & 0.5929 & 0.4247 & 0.8278 & 0.1722 \\
    GBDT & None & RUS & 0.7294 & 0.7087 & 0.5887 & 0.4195 & 0.8261 & 0.1739 \\
    GBDT & None & IHT & 0.7283 & 0.7062 & 0.5869 & 0.4173 & 0.8255 & 0.1745 \\
    GBDT & None & ENN & 0.7292 & 0.7079 & 0.5883 & 0.4191 & 0.8260 & 0.1740 \\
    GBDT & None & ROS & 0.7303 & 0.7085 & 0.5903 & 0.4214 & 0.8268 & 0.1732 \\
    GBDT & None & SMOTE & 0.7246 & 0.7013 & 0.5804 & 0.4094 & 0.8230 & 0.1770 \\
    GBDT & None & ADASYN & 0.7237 & 0.7005 & 0.5789 & 0.4076 & 0.8224 & 0.1776 \\
    GBDT & None & SMOTEENN & 0.7268 & 0.7010 & 0.5843 & 0.4141 & 0.8245 & 0.1755 \\
    GBDT & None & SMOTETomek & 0.7240 & 0.7015 & 0.5795 & 0.4082 & 0.8226 & 0.1774 \\
    GBDT & COPOD & None & 0.7312 & 0.7105 & 0.5918 & 0.4233 & 0.8273 & 0.1727 \\
    GBDT & COPOD & RUS & 0.7289 & 0.7076 & 0.5879 & 0.4185 & 0.8258 & 0.1742 \\
    GBDT & COPOD & IHT & 0.7291 & 0.7056 & 0.5882 & 0.4189 & 0.8260 & 0.1740 \\
    GBDT & COPOD & ENN & 0.7306 & 0.7078 & 0.5908 & 0.4221 & 0.8270 & 0.1730 \\
    GBDT & COPOD & ROS & \textbf{0.7327} & 0.7097 & \textbf{0.5942} & \textbf{0.4264} & \textbf{0.8283} & \textbf{0.1717} \\
    GBDT & COPOD & SMOTE & 0.7247 & 0.7002 & 0.5805 & 0.4095 & 0.8230 & 0.1770 \\
    GBDT & COPOD & ADASYN & 0.7237 & 0.6996 & 0.5789 & 0.4076 & 0.8224 & 0.1776 \\
    GBDT & COPOD & SMOTEENN & 0.7272 & 0.7024 & 0.5849 & 0.4148 & 0.8247 & 0.1753 \\
    GBDT & COPOD & SMOTETomek & 0.7243 & 0.7004 & 0.5799 & 0.4087 & 0.8228 & 0.1772 \\
    GBDT & IForest & None & 0.7315 & 0.7106 & 0.5923 & 0.4239 & 0.8275 & 0.1725 \\
    GBDT & IForest & RUS & 0.7301 & 0.7096 & 0.5899 & 0.4210 & 0.8266 & 0.1734 \\
    GBDT & IForest & IHT & 0.7284 & 0.7053 & 0.5870 & 0.4174 & 0.8255 & 0.1745 \\
    GBDT & IForest & ENN & 0.7300 & 0.7078 & 0.5897 & 0.4208 & 0.8265 & 0.1735 \\
    GBDT & IForest & ROS & 0.7300 & 0.7094 & 0.5897 & 0.4208 & 0.8265 & 0.1735 \\
    GBDT & IForest & SMOTE & 0.7244 & 0.7017 & 0.5801 & 0.4090 & 0.8229 & 0.1771 \\
    GBDT & IForest & ADASYN & 0.7238 & 0.6996 & 0.5791 & 0.4077 & 0.8225 & 0.1775 \\
    GBDT & IForest & SMOTEENN & 0.7261 & 0.7017 & 0.5830 & 0.4126 & 0.8240 & 0.1760 \\
    GBDT & IForest & SMOTETomek & 0.7235 & 0.7008 & 0.5785 & 0.4071 & 0.8223 & 0.1777 \\
    GBDT & CBLOF & None & 0.7309 & 0.7104 & 0.5912 & 0.4226 & 0.8271 & 0.1729 \\
    GBDT & CBLOF & RUS & 0.7295 & 0.7088 & 0.5888 & 0.4197 & 0.8262 & 0.1738 \\
    GBDT & CBLOF & IHT & 0.7299 & 0.7055 & 0.5895 & 0.4205 & 0.8265 & 0.1735 \\
    GBDT & CBLOF & ENN & 0.7293 & 0.7077 & 0.5885 & 0.4193 & 0.8261 & 0.1739 \\
    GBDT & CBLOF & ROS & 0.7311 & 0.7090 & 0.5916 & 0.4231 & 0.8273 & 0.1727 \\
    GBDT & CBLOF & SMOTE & 0.7243 & 0.7017 & 0.5799 & 0.4088 & 0.8228 & 0.1772 \\
    GBDT & CBLOF & ADASYN & 0.7237 & 0.6991 & 0.5789 & 0.4076 & 0.8224 & 0.1776 \\
    GBDT & CBLOF & SMOTEENN & 0.7271 & 0.7026 & 0.5847 & 0.4146 & 0.8246 & 0.1754 \\
    GBDT & CBLOF & SMOTETomek & 0.7245 & 0.7018 & 0.5802 & 0.4092 & 0.8229 & 0.1771 \\
    GBDT & LOF & None & 0.7310 & \textbf{0.7107} & 0.5913 & 0.4228 & 0.8272 & 0.1728 \\
    GBDT & LOF & RUS & 0.7300 & 0.7081 & 0.5897 & 0.4208 & 0.8265 & 0.1735 \\
    GBDT & LOF & IHT & 0.7280 & 0.7057 & 0.5863 & 0.4166 & 0.8252 & 0.1748 \\
    GBDT & LOF & ENN & 0.7305 & 0.7085 & 0.5906 & 0.4219 & 0.8269 & 0.1731 \\
    GBDT & LOF & ROS & 0.7304 & 0.7099 & 0.5903 & 0.4215 & 0.8268 & 0.1732 \\
    GBDT & LOF & SMOTE & 0.7247 & 0.7010 & 0.5806 & 0.4096 & 0.8231 & 0.1769 \\
    GBDT & LOF & ADASYN & 0.7237 & 0.7001 & 0.5788 & 0.4075 & 0.8224 & 0.1776 \\
    GBDT & LOF & SMOTEENN & 0.7265 & 0.7017 & 0.5836 & 0.4133 & 0.8242 & 0.1758 \\
    GBDT & LOF & SMOTETomek & 0.7247 & 0.7011 & 0.5806 & 0.4096 & 0.8231 & 0.1769 \\ \hline
    \end{tabular}
\end{table}

\subsection{Comprehensive comparison of machine learning models}
Table \ref{table: comparison of models} synthesizes the prediction results of the four models. GBDT obtains the best results whether comparing the case without outlier detection and balanced sampling method (AUC 0.7102), comparing the average values of all same-model methods (9*5=45 types, AUC 0.7051), or comparing the best values that the models can obtain (AUC 0.7107).

Among the best strategies for each model, LOF outlier detection and no sample balance methods each take three seats. IForest is more effective for RF, and IHT downsampling methods are more effective for MLP. RF performs better than MLP when outlier detection and balanced sampling methods are not used. However, combining LOF and IHT, the AUC of the best MLP model even surpassed the best RF model.

\begin{table}[H]\footnotesize
    \centering
    \caption{Comprehensive comparison of the machine learning models.}
    \label{table: comparison of models}
    \begin{tabular}{llllllllll}
    \hline
    Mode & Model & Reject strategy & Sampling strategy & ACC & AUC & G-mean & TPR & TNR & FPR \\ \hline
    \multirow{4}{*}{None} & GBDT & None & None & \textbf{0.7319} & \textbf{0.7102} & \textbf{0.5929} & \textbf{0.4247} & \textbf{0.8278} & \textbf{0.1722} \\
     & LR & None & None & 0.7298 & 0.7071 & 0.5893 & 0.4203 & 0.8264 & 0.1736 \\
     & RF & None & None & 0.7242 & 0.6977 & 0.5797 & 0.4085 & 0.8227 & 0.1773 \\
     & MLP & None & None & 0.7230 & 0.6953 & 0.5778 & 0.4061 & 0.8220 & 0.1780 \\ \hline
    \multirow{4}{*}{Mean} & GBDT & All mean & All mean & 0.7277 & \textbf{0.7051} & 0.5858 & 0.4160 & 0.8251 & 0.1749 \\
     & LR & All mean & All mean & \textbf{0.7287} & 0.7046 & \textbf{0.5875} & \textbf{0.4181} & \textbf{0.8257} & \textbf{0.1743} \\
     & RF & All mean & All mean & 0.7192 & 0.6895 & 0.5712 & 0.3982 & 0.8195 & 0.1805 \\
     & MLP & All mean & All mean & 0.7068 & 0.6574 & 0.5490 & 0.3721 & 0.8113 & 0.1887 \\ \hline
    \multirow{4}{*}{Best} & GBDT & LOF & None & \textbf{0.7310} & \textbf{0.7107} & \textbf{0.5913} & \textbf{0.4228} & \textbf{0.8272} & \textbf{0.1728} \\
     & LR & LOF & None & 0.7299 & 0.7079 & 0.5895 & 0.4205 & 0.8265 & 0.1735 \\
     & RF & IForest & None & 0.7248 & 0.6985 & 0.5808 & 0.4098 & 0.8231 & 0.1769 \\
     & MLP & LOF & IHT & 0.7284 & 0.7041 & 0.5870 & 0.4174 & 0.8255 & 0.1745 \\ \hline
    \end{tabular}
    \begin{tablenotes}
        \item The mode ``None'' is the case comparing the models' classification results without outlier detection and balanced sampling methods. The mode ``Mean'' is the case comparing the average classification results of each model using every outlier detection and balanced sampling methods. The mode ``Best'' is the case comparing the best classification results that the models can achieved.
    \end{tablenotes}
\end{table}

\subsection{Comprehensive comparison of the outlier detection methods}
Table \ref{table: comparison of outlier detection} synthesizes the results of the different outlier detection methods. In general, as long as the outlier detection methods are used, the effect will be better than without it. After taking the average of the experimental results of the same outlier detection algorithms (9*4=36 types), the ranking from highest to lowest is IForest, COPOD, CBLOF, LOF, None. The average classification results using IForest is the best, but in most cases, LOF can help the model get the best results according to table \ref{table: comparison of models}.

\begin{table}[H]\footnotesize
    \centering
    \caption{Comprehensive comparison of the outlier detection methods.}
    \label{table: comparison of outlier detection}
    \begin{tabular}{llllllll}
    \hline
    Reject strategy & ACC & AUC & G-mean & TPR & TNR & FPR & Rank \\ \hline
    IForest & 0.7207 & \textbf{0.6896} & 0.5734 & 0.4011 & 0.8204 & 0.1796 & \textbf{1.5} \\
    COPOD & \textbf{0.7208} & 0.6891 & \textbf{0.5737} & \textbf{0.4014} & \textbf{0.8205} & \textbf{0.1795} & 3 \\
    CBLOF & 0.7204 & 0.6891 & 0.5731 & 0.4007 & 0.8203 & 0.1797 & 3 \\
    LOF & 0.7207 & 0.6891 & 0.5735 & 0.4012 & 0.8204 & 0.1796 & 3.25 \\
    None & 0.7206 & 0.6888 & 0.5733 & 0.4010 & 0.8204 & 0.1796 & 4.25 \\ \hline
    \end{tabular}
    \begin{tablenotes}
        \item The ``Rank'' is calculated as the mean of the ranking (AUC) of outlier detection methods under the same model and the same balanced sampling method.
    \end{tablenotes}
\end{table}

\subsection{Comprehensive comparison of the balanced sampling methods}
Table \ref{table: comparison of balanced sampling} synthesizes the results of the different balanced sampling methods. After taking the mean of the experimental results of the same balanced sampling methods (5*4=20 types), it is found that the effect of not using a balanced sampling method is the best, simple RUS and ROS are ranked 2, 3. While complex balanced sampling methods are less effective, and the commonly used SMOTE algorithm is ranked only the second to last. The ADASYN algorithm, which achieved the worst results among the four types of models, ranked first from the bottom. Except for the effect of IHT on MLP, for other models, the balanced sampling method will reduce the classification results. The ranking is similar to two relevant studies \citep{Moscato2021, Namvar2018} which using the Lending Club data too. In their research, RUS also achieved the best results among all sampling methods. However, neither of them added the no sampling method to the experiments as a comparison.

\begin{table}[H]\footnotesize
    \centering
    \caption{Comprehensive comparison of the different balanced sampling methods}
    \label{table: comparison of balanced sampling}
    \begin{tabular}{llllllll}
    \hline
    Sampling strategy & ACC & AUC & G-mean & TPR & TNR & FPR & Rank \\ \hline
    None & \textbf{0.7274} & \textbf{0.7029} & \textbf{0.5853} & \textbf{0.4154} & \textbf{0.8249} & \textbf{0.1751} & \textbf{1.25} \\
    RUS & 0.7264 & 0.7018 & 0.5836 & 0.4133 & 0.8242 & 0.1758 & 2.75 \\
    ROS & 0.7187 & 0.6853 & 0.5699 & 0.3971 & 0.8191 & 0.1809 & 3.25 \\
    ENN & 0.7233 & 0.6941 & 0.5781 & 0.4066 & 0.8221 & 0.1779 & 4 \\
    IHT & 0.7259 & 0.7008 & 0.5826 & 0.4121 & 0.8238 & 0.1762 & 4.5 \\
    SMOTEENN & 0.7193 & 0.6864 & 0.5713 & 0.3984 & 0.8195 & 0.1805 & 6.25 \\
    SMOTETomek & 0.7155 & 0.6782 & 0.5643 & 0.3902 & 0.8170 & 0.1830 & 6.75 \\
    SMOTE & 0.7155 & 0.6779 & 0.5644 & 0.3904 & 0.8171 & 0.1829 & 7.25 \\
    ADASYN & 0.7137 & 0.6750 & 0.5611 & 0.3864 & 0.8158 & 0.1842 & 9 \\ \hline
    \end{tabular}
    \begin{tablenotes}
        \item The ``Rank'' is calculated as the mean of the ranking (AUC) of balanced sampling methods in the case of the same model and the same outlier detection method.
    \end{tablenotes}
\end{table}

\section{Conclusions}
\label{conclusions}
Credit scoring is crucial for P2P lending platforms, outliers and sample imbalance make the construction of credit scoring models challenging. There have been many studies using balanced sampling methods to deal with the sample imbalance problem \citep{Moscato2021, Namvar2018, Niu2020}, but few studies have incorporated outlier detection methods into the modeling process.

This paper provides a systematic study for the application of outlier detection and balanced sampling methods in the field of credit scoring. The experiments use real P2P lending platform data and cover a variety of outlier detection, balanced sampling methods, and machine learning models. The experimental results show that the outlier detection algorithm can enhance the robustness of the model. In addition, a suitable balanced sampling method brings a large improvement to MLP but is relatively less useful to LR, RF, and GBDT.

In future plans, firstly, more advanced models, such as ensemble strategies or deep learning models, will be studied. Secondly, feature selection is also an important part of the construction of the credit scoring model, which will be considered to be integrated into the existing framework.

\section*{Acknowledgments}
This work was supported by HuaRong RongTong (Beijing) Technology Co., Ltd. We acknowledge HuaRong RongTong (Beijing) for providing us with high-performance machines for computation. We also acknowledge the anonymous reviewers for proposing detailed modification advice to help us improve the quality of this manuscript.

\bibliography{sample}

\end{document}